# The Use of AI Tools to Develop and Validate Q-Matrices

Kevin Fan, Jacquelyn A. Bialo, and Hongli Li[†*]

†Department of Educational Policy Studies, Georgia State University, Atlanta, GA. USA.
*Corresponding author. Email: hli24@gsu.edu

**Abstract**

Constructing a Q-matrix is a critical but labor-intensive step in cognitive diagnostic modeling (CDM). This study investigates whether AI tools (i.e., general language models) can support Q-matrix development by comparing AI-generated Q-matrices with a validated Q-matrix from Li and Suen (2013) for a reading comprehension test. In May 2025, multiple AI models were provided with the same training materials as human experts. Agreement among AI-generated Q-matrices, the validated Q-matrix, and human raters' Q-matrices was assessed using Cohen's kappa. Results showed substantial variation across AI models, with Google Gemini 2.5 Pro achieving the highest agreement ($\kappa$ = 0.63) with the validated Q-matrix, exceeding that of all human experts. A follow-up analysis in January 2026 using newer AI versions, however, revealed lower agreement with the validated Q-matrix. Implications and directions for future research are discussed.

## 1. Background

Cognitive diagnostic modeling (CDM) is a valuable approach that allows assessments to provide rich diagnostic information to inform instruction and learning (Rupp & Templin, 2008). To enable the use of CDM, a critical first step is to map test items onto an item-by-skill table, known as a Q-matrix (Tatsuoka, 1983). A Q-matrix represents a particular hypothesis about which skills (or attributes) are required to successfully answer each item in a test. A correct answer to the item may depend on one or more skills. Developing a Q-matrix is an iterative process that requires both theoretical guidance and content knowledge about the construct being tested. After an initial set of attributes has been specified, the Q-matrix needs to be refined to ensure that there are sufficient high-quality items for each attribute to produce stable results. Therefore, developing a Q-matrix, particularly for existing assessments, can be both challenging and time-consuming (Hunter et al., 2022).

While generative AI is increasingly recognized for assisting with different stages of educational measurement, such as test development, automatic scoring, and feedback giving (Bulut et al., 2024), its capacity to facilitate building a Q-matrix remains understudied. Aşiret and Sünbül (2025) evaluated ChatGPT 4o's performance in building a Q-matrix for a short math test consisting of 12 fraction subtraction items. The Q-matrix for this dataset included four skills: (a) performing basic fraction subtraction operations, (b) simplified/reduced fractions, (c) separating whole numbers from fractions, and (d) borrowing from whole numbers to fractions. The authors found that the Q-matrix generated by ChatGPT 4o had a high overlap with those constructed by researchers and human experts. Although these findings are encouraging, the subtraction items examined were relatively straightforward. In contrast, constructing a Q-matrix for a reading comprehension assessment involves more complex and abstract cognitive processes, making it substantially more challenging for automated approaches.

The purpose of this study is, therefore, to examine the application of a series of AI tools in Q-matrix development and validation of a reading comprehension test across time. Specifically, it compares the performance of several AI tools in generating Q-matrices for a reading comprehension test originally used in Li and Suen (2013).

## 2. Methods

### *2.1 Original Q-Matrix in Li and Suen (2013)*

The reading test examined in Li and Suen (2013) includes four passages, each followed by five multiple-choice items. The underlying reading comprehension construct consists of two broad categories: linguistic skills and comprehension skills. Linguistic skills include vocabulary (Skill 1) and syntax (Skill 2), whereas comprehension skills include extracting explicit information (Skill 3), connecting and synthesizing (Skill 4), and making inferences (Skill 5). A Q-matrix was developed based on student think-aloud protocols and expert judgment, then validated with empirical data. Their validated Q-matrix is presented in Table 1.

Four experts were invited to identify the skills required by each of the 20 items in the reading comprehension test. All experts were advanced doctoral students in education or applied linguistics with extensive experience teaching English reading to English as a Second Language learners. To familiarize the experts with the rating task, a training session was conducted prior to formal coding. During this session, the reading test, the cognitive model of reading, and the associated coding scheme were introduced. Following training, each expert independently read the passages and completed the rating task in a library study room. Experts identified the skills required for each item and documented annotations describing the evidence supporting their judgments.

**Table 1.** Validated Q-Matrix in Li and Suen (2013)

| Item | Skill 1 (Vocabulary) | Skill 2 (Syntax) | Skill 3 (Extracting explicit information) | Skill 4 (Connecting and synthesizing) | Skill 5 (Making inferences) |
|---|---|---|---|---|---|
| 1  | 1 | 0 | 0 | 1 | 0 |
| 2  | 1 | 0 | 0 | 0 | 0 |
| 3  | 0 | 0 | 0 | 1 | 0 |
| 4  | 0 | 0 | 1 | 0 | 0 |
| 5  | 1 | 0 | 0 | 0 | 1 |
| 6  | 1 | 0 | 1 | 0 | 0 |
| 7  | 0 | 0 | 1 | 0 | 0 |
| 8  | 1 | 0 | 0 | 1 | 0 |
| 9  | 0 | 0 | 1 | 0 | 0 |
| 10 | 1 | 0 | 0 | 0 | 1 |
| 11 | 0 | 0 | 1 | 0 | 0 |
| 12 | 1 | 1 | 0 | 0 | 0 |
| 13 | 0 | 0 | 0 | 1 | 0 |
| 14 | 1 | 0 | 0 | 1 | 0 |
| 15 | 1 | 1 | 0 | 0 | 1 |
| 16 | 1 | 1 | 1 | 0 | 0 |
| 17 | 0 | 1 | 0 | 1 | 0 |
| 18 | 0 | 1 | 1 | 0 | 0 |
| 19 | 1 | 0 | 0 | 1 | 0 |
| 20 | 0 | 0 | 1 | 0 | 0 |

### *2.2 The Use of AI Tools to Build Q-matrix*

In the present study, AI models were asked to follow the same procedures as the human experts and were provided with identical training materials. In May 2025, the training materials were supplied to several AI models, including ChatGPT (versions 4o and 4.5), Google Gemini 2.5 Pro, Claude 4 Sonnet, and DeepSeek V3. The AI-generated Q-matrices were compared with the validated Q-matrix reported in Li and Suen (2013) using

Cohen's kappa. For benchmarking purposes, the Q-matrices produced by human experts were also compared with the validated matrix. The results of the first round were presented at the Psychometric Society Conference in July 2025 (Fan et al., 2025).

Because the AI models had been updated since the original results were presented, in January 2026, the same training materials were provided to newer versions of these models, including ChatGPT 5.2, Google Gemini 3 Pro, Claude 4.5 Sonnet, and DeepSeek V3.2. It is generally understood that the newer AI models incorporate advances in general reasoning, semantic abstraction, and long-context processing relative to earlier versions. The results from the second round of analyses were also briefly summarized.

The prompt used for both rounds was: *"For the rating task, you will be provided with 4 short reading passages, each followed by 5 multiple-choice items. You will be asked to identify what reading skills are required to answer each item correctly. If you believe the skill is required by the item, please put a 1 in the table below, otherwise, please put a zero. Please also provide justifications for your decisions."*

## 3. Results and Discussion

### 3.1 Primary Analysis from May 2025

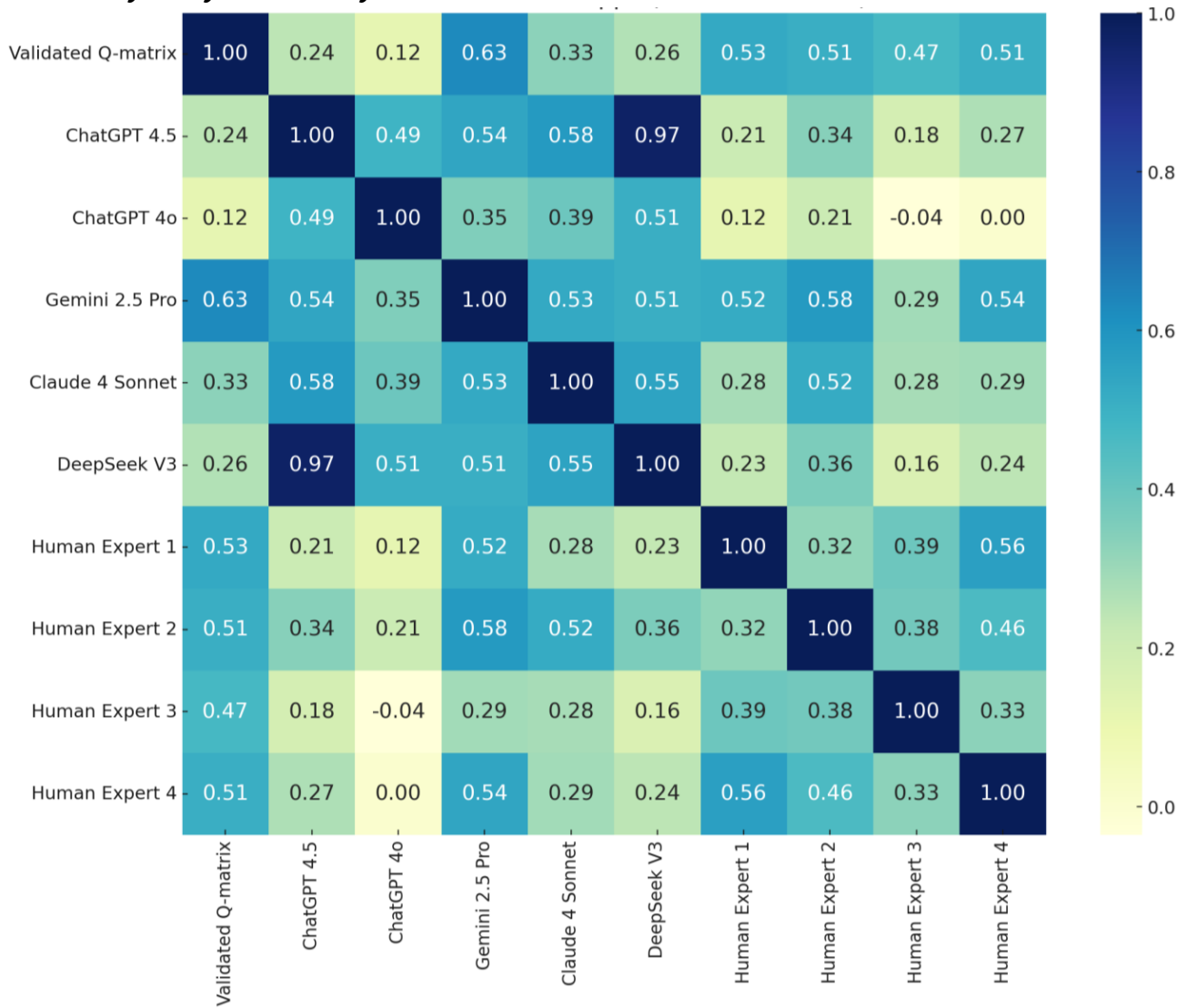

**Figure 1.** Cohen's Kappa Comparing AI-Generated Q-matrices, Human Raters' Q-Matrices, and the Validated Q-Matrix (May 2025).

Figure 1 shows the pairwise Cohen's Kappa value comparing AI-Generated and Human Raters' Q-Matrices with the Validated Q-Matrix. Following the guidelines of Altman (1991), adapted from Landis and Koch (1977), values below 0.20 indicate poor agreement; 0.21–0.40, fair; 0.41–0.60, moderate; 0.61–0.80, good; and 0.81–1.00, very good agreement.

Gemini 2.5 Pro showed the highest agreement with the validated Q-matrix ($k$ = 0.63), followed by Claude 4 Sonnet ($k$ = 0.33), DeepSeek V3 ($k$ = 0.26), ChatGPT 4.5 ($k$ = 0.24), and ChatGPT 4o ($k$ = 0.12). Human experts generally showed slightly higher agreement with the validated Q-matrix than most AI tools. However, it is important to note that Gemini 2.5 Pro outperformed all human experts particularly in identifying comprehension skills (extracting explicit information, connecting and synthesizing, and making inferences). In addition, ChatGPT 4.5 and DeepSeek V3 demonstrated very high agreement with each other ($k$ = 0.97), suggesting strong internal consistency between these two, but lower alignment with the validated Q-matrix.

Figure 2 further illustrates Cohen's Kappa values across skill categories. Both human experts and AI tools showed better accuracy in coding comprehension skills (extracting explicit information, connecting and synthesizing, and making inferences) than linguistic skills (vocabulary and syntax). This pattern was likely due to the complexity involved in understanding the linguistic skills and limitations in the original coding framework.

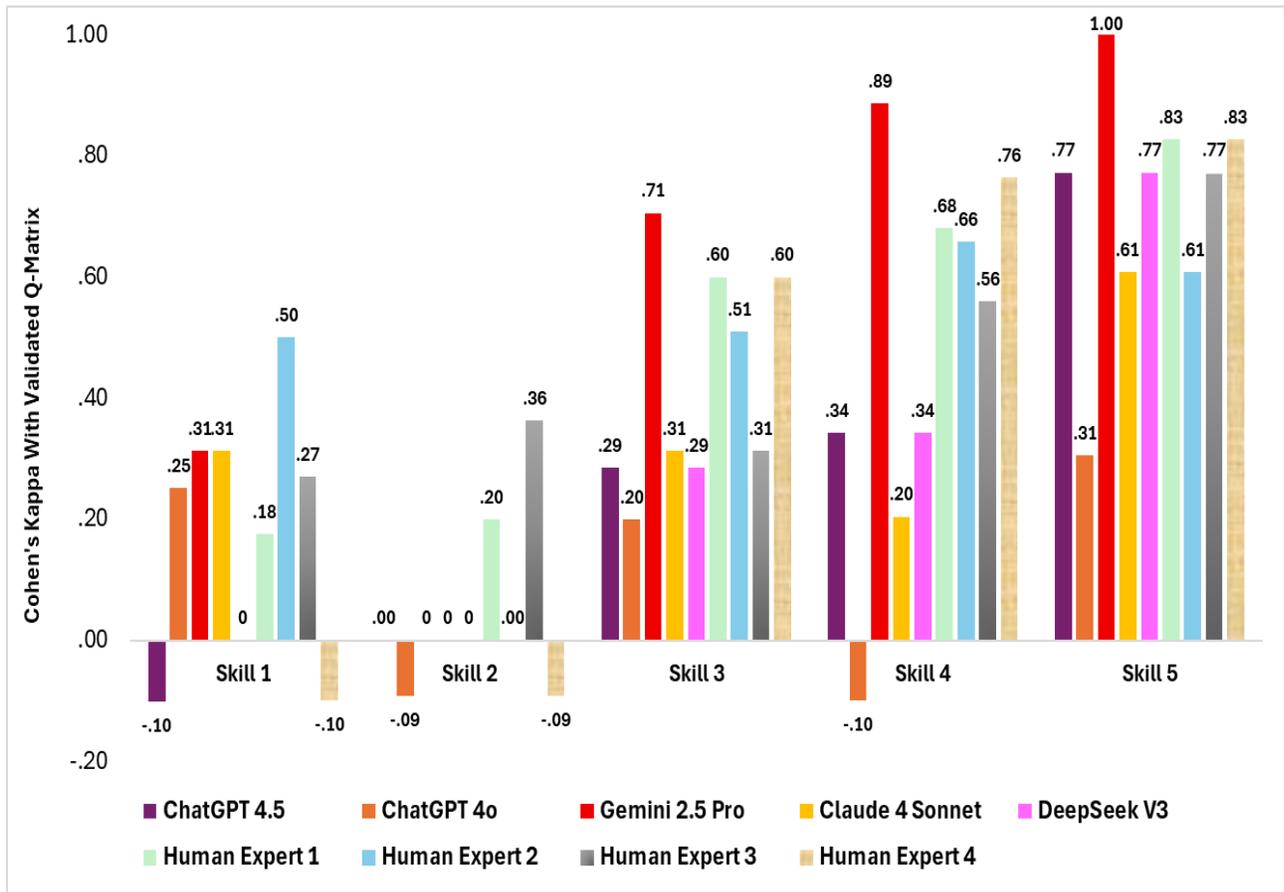

**Figure 2.** Cohen's Kappa Comparing AI-Generated and Human Raters' Q-Matrices with the Validated Q-Matrix Across Skills (May 2025).

### *3.2 Additional Analysis from January 2026*

As shown in Figure 3, among the newer AI models, Google Gemini 3 Pro still showed the highest agreement with the validated Q-matrix ($\kappa$ = 0.42), followed by Claude 4.5 Sonnet ($\kappa$ = 0.26), DeepSeek V3.2 ($\kappa$ = 0.21), and ChatGPT 5.2 ($\kappa$ = 0.16). Although Gemini 3 Pro is still the best performing model, its agreement was much lower than that of the earlier Gemini 2 Pro ($k$ = 0.63). Overall, agreement levels for the newer AI models were lower than those observed for earlier versions in May 2025, and all the newer models showed worse performance than all four human raters.

Upon close examination, we found that ChatGPT 5.2 did not code linguistic skills (vocabulary and syntax) for any items. The model justified this decision by stating that "vocabulary and syntax were not coded because no item hinges on decoding infrequent words or complex sentence structures." In contrast, earlier versions (ChatGPT 4.5 and ChatGPT 4o) coded linguistic skills for a subset of items. This discrepancy suggests that, although the newer model demonstrates enhanced reasoning capability, it may reinterpret task requirements and deviate from the provided coding rules. This pattern is consistent with prior literature indicating that instruction-tuned large language models are encouraged to reason about and reinterpret rules rather than adhere to them rigidly (Bai et al., 2022). Therefore, the findings from this study suggest that improvements in AI models' general reasoning ability do not necessarily translate into closer alignment with the validated Q-matrix structure.

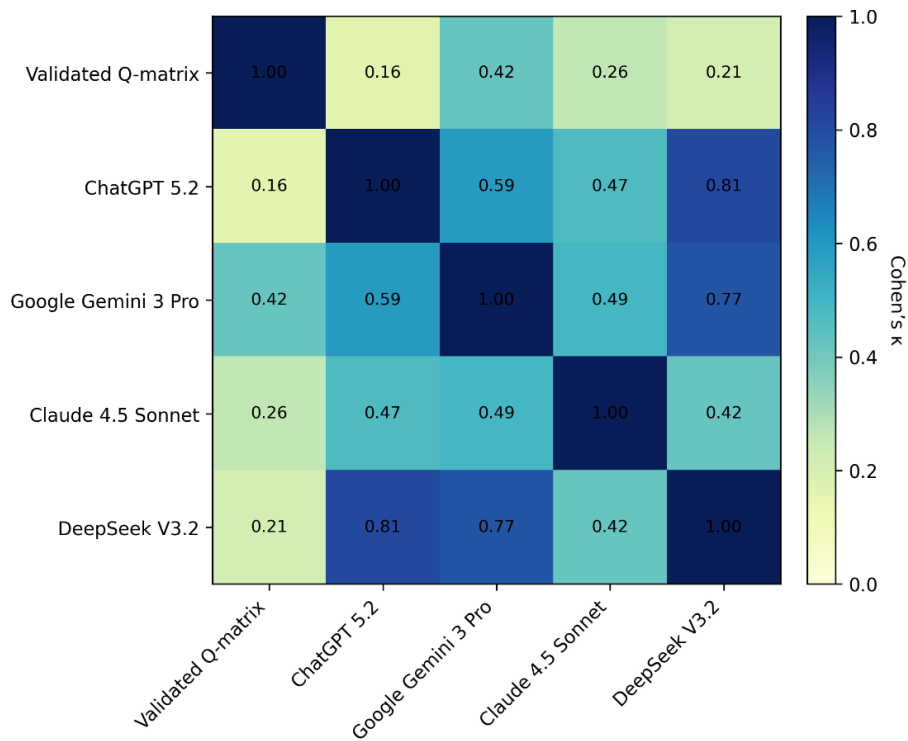

**Figure 3.** Cohen's Kappa Between Newer AI-Generated Q-Matrices and the Validated Q-Matrix (January 2026).

## 4. Conclusion

Constructing a Q-matrix is a critical but labor-intensive step in CDM, requiring substantial theoretical grounding and expert judgment. Recent advances in generative AI suggest potential for supporting this process, yet empirical evidence remains limited. This study examines the extent to which AI tools can generate Q-matrices that align with a validated Q-matrix for a reading comprehension test originally reported in Li and Suen (2013). Results from the primary analysis conducted in May 2025 indicate that AI tools can produce Q-matrices that show meaningful agreement with the validated Q-matrix, although performance varies substantially across models. Notably, Google Gemini 2.5 Pro demonstrated the highest agreement with the validated Q-matrix, outperforming all individual human experts, particularly in identifying comprehension-related skills. However, when newer versions of AI models were used in January 2026 for the same task, the AI generated Q-matrices showed lower agreement with the validated Q-matrix, though Google Gemini 3 Pro was still the best performing one.

In summary, AI models show some potential to facilitate the Q-matrix construction process. However, we observed substantial variability across different AI tools and time points. Although Gemini 2.5 Pro demonstrated better performance than the four human raters, its newer version Gemini 3 Pro in fact performed worse than the human raters. These findings suggest that AI tools may be best used as a supplementary source of evidence to support Q-matrix development, rather than as a replacement for expert judgment. As such, human expert judgment remains a critical component of Q-matrix construction for CDM analyses.

This study has some limitations. First, it relied on a single reading comprehension assessment, which limits the generalizability of the findings. Second, the reading comprehension construct itself is complex and multifaceted, and any single cognitive framework may provide an incomplete representation of the underlying skills. Third, some uncertainty may exist regarding the originally validated Q-matrix used as the benchmark. In addition, AI-generated results may vary across runs due to the probabilistic nature of large language models, whereas the present study relied on a zero-shot approach. Future research should examine different reading assessments and cognitive frameworks, as well as explore alternative prompting strategies and training approaches to improve the consistency and accuracy of AI-assisted Q-matrix construction.